\documentclass[runningheads,orivec]{llncs}

\usepackage[T1]{fontenc}

\usepackage[hyphens]{url}  
\usepackage{graphicx} 
\usepackage{amsmath}  
\usepackage{booktabs}

\usepackage{algorithm}
\usepackage{algorithmic}
\usepackage{subcaption}
\usepackage[capitalize,noabbrev]{cleveref}

\usepackage{xspace}
\newcommand{\rex}{\textsc{ReX}\xspace}
\newcommand{\specrex}{\textsc{SpecReX}\xspace}
\newcommand{\xai}{\textsc{XAI}\xspace}

\begin{document}
\title{SpecReX: Explainable AI for Raman Spectroscopy}

\author{
Nathan Blake\inst{1,2}\orcidID{0000-0002-6404-514X} \and 
David A. Kelly\inst{2}\orcidID{0000-0002-5368-6769} \and
Akchunya Chanchal\inst{2}\orcidID{0000-0003-2571-0802} \and 
Sarah Kapllani-Mucaj\inst{1}\orcidID{0000-0003-3238-4224} \and 
Geraint Thomas\inst{1}\orcidID{0000-0003-4035-7832} \and 
Hana Chockler\inst{2}\orcidID{0000-0003-1219-0713}
}

\authorrunning{N. Blake et al.}

\institute{
University College London, UK \and 
King's College London, UK
\email{\{nathan.blake,david.a.kelly,akchunya.chanchal,hana.chockler\}@kcl.ac.uk}
\email{\{sarah.mucaj.23,g.thomas\}@ucl.ac.uk}
}

\maketitle

\begin{abstract}
Raman spectroscopy is becoming more common for medical diagnostics with deep learning models being increasingly used to leverage its full potential. However, the opaque nature of such models and the sensitivity of medical diagnosis together with regulatory requirements necessitate the need for explainable AI tools. We introduce \specrex, specifically adapted to explaining Raman spectra. \specrex uses the theory of actual causality to rank causal responsibility in a spectrum, quantified by iteratively refining mutated versions of the spectrum and testing if it retains the original classification. The explanations provided by \specrex take the form of a responsibility map, highlighting spectral regions most responsible for the model to make a correct classification. To assess the validity of \specrex, we create increasingly complex simulated spectra, in which a ‘ground truth’ signal is seeded, to train a classifier. We then obtain \specrex explanations and compare the results with another explainability tool. By using simulated spectra we establish that \specrex localizes to the known differences between classes, under a number of conditions. This provides a foundation on which we can find the spectral features which differentiate disease classes. This is an important first step in proving the validity of \specrex.

\end{abstract}

\section{Introduction}

The deep learning (DL) paradigm holds much promise to enhance many aspects of healthcare, particularly to improve and expedite medical diagnostics~\cite{esteva2019guide}. This potential extends to biophotonics \cite{pradhan2020deep} and research in the domain of biomedical Raman spectroscopy (RS), which is increasingly moving away from traditional machine learning (ML) paradigms towards DL \cite{blake2022machine}. This is due to DL being able to capture complex, non-linear relationships between inputs without needing to perform feature selection. This is important, as biomedical samples, such as tissue biopsies, are themselves highly complex and differences between disease classes are incredibly subtle. However, DL is notoriously difficult to interpret, particularly when compared to some forms of traditional ML. There is typically an inverse relation between the performance of a model and its interpretability~\cite{luo2019balancing}, with less interpretable models being described as 'black-box' models. Adding to that, many models, especially in healthcare, are
proprietary, making them black-box regardless of their complexity.
Helping to shine a light upon these opaque models, eXplainable AI (\xai) is a fast developing field in computer science. 

In this paper, we introduce a causal explainability algorithm for DL models trained on RS data. We introduce an implementation of our method, \specrex, which is an extension of the causal \xai tool \rex~\cite{chockler2024causal}. Given that real spectral data is extremely complex and noisy, we design an experiment whereby we use synthetic data, so that we know the ground truth of the signal. We compare the explanations provided by \specrex, modulo a model trained on similar synthetic data, against the ground truth. 
\specrex is part of the \rex framework and is available at \url{https://github.com/ReX-XAI/ReX}.

\section{Background and Related Work}\label{sec:relwork}

\begin{figure}[t]
\centering
   \includegraphics[width=\columnwidth]{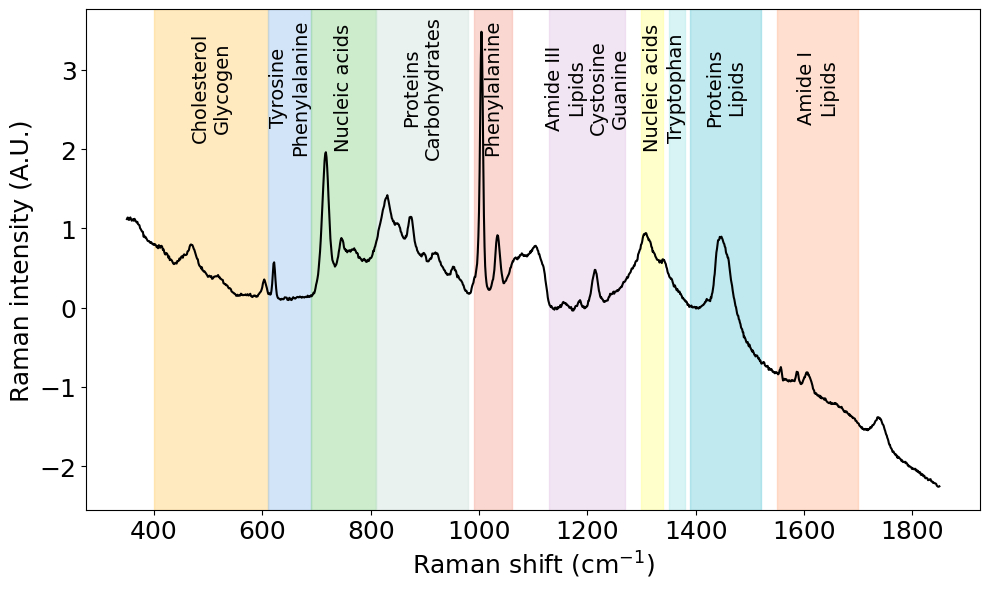}
   \caption{A typical biological Raman spectrum. The x-axis shows the extent of wavenumber, or Raman, shift. The y-axis shows the number of photons, also called the intensity - this is often normalized and so reported in arbitrary units (A.U.) The colored regions approximately correspond to various biochemical features of interest to a clinical problem. Figure adjusted from \cite{delrue2022potential}.}
  \label{fig:typicalSpectrum}
\end{figure}

Raman spectroscopy is a vibrational spectroscopic technique which interrogates the vibrational modes of a molecule. When photons interact with the vibrational modes of a molecule, a certain proportion of photons will scatter inelastically: they will gain or lose some energy. Such scattering is called \emph{Raman scattering}. If a monochromatic light source, such as a laser, is shone upon an a molecule, it is possible to measure the precise amount by which Raman scattered photons have gained or lost energy (known as stokes and anti-stokes Raman scattering respectively). A Raman spectrum shows the number of photons (y-axis) which have had their wavenumber shifted by a certain amount, measured by reciprocal centimeters  $cm^{-1}$ (x-axis). Each unique vibrational mode inelastically scatters in a unique manner, allowing Raman spectra to be interpreted in terms of the underlying chemistry of the sample (\Cref{fig:typicalSpectrum}).

The technology has been increasingly explored for its applicability to various biomedical diagnostics \cite{butler2016using}. However, the complexity of biological samples means that the application of traditional ML is insufficient for the precision required for medical applications, such as oncology \cite{blake2022machine}. For this reason, DL architectures are becoming more popular and so too is interest in \xai for RS. 

There are no explicit methods for explainability particular to RS. However, there are various attempts to extract meaning from the results of ML-based RS studies. A commonly-used method is obtaining the mean spectra of pertinent disease classes and then comparing the differences. The spectral features thus highlighted by the difference between two mean spectra are then assumed to be responsible for a ML model's classification. Another method is to examine the outputs of preprocessing transformations before ML. For instance, Principal Component Analysis (PCA) is often used as a dimensionality reduction technique. The loadings of the resultant principal components can be manually inspected for spectral features common to a biomolecular entity. However, in both cases, it does not necessarily follow that any DL model is using these highlighted features. For instance, there may be spurious differences between spectra which a DL model learns to ignore (assuming it has not been overfitted). Even if the differences are real, it does not necessarily follow that they are the ones which the model is using to distinguish between disease classes.

There have been a few attempts to apply \xai to biomedical RS. A very simple occlusion-based method attempted to identify spectral features responsible for distinguishing between sub-types of colorectal cancer in human tissues based on a convolutional neural network (CNN) \cite{blake2023deep}. Two papers have applied SHAP (SHapley Additive exPlanations) \cite{lundberg2017unified}, which was developed for general use on an array of different data types. The first paper applied SHAP to Raman peak ratios in order to identify spectral features utilized in a binary classification of thyroid tissues  \cite{bellantuono2023explainable}, and the second applied it to identify responsible Raman peaks in a melanoma classifier \cite{ibtehaz2023ramannet}.
Gradient based \xai methods have also been applied to find spectral regions of importance for distinguishing between the Raman spectra of cancerous and healthy DNA, due to methylation \cite{lancia2023learning}. None of these methods take into account the unique nature of spectral data. In addition,  none have been rigorously tested for their validity upon Raman spectra. Such rigorous testing is required for medical applications of \xai, given the critical nature of decisions which may be made based in their outputs. 

Jin \textit{et al}. \cite{jin2023guidelines} identify four key criteria on which to assess medically deployed \xai tools: understandability, clinical relevance, truthfulness and informative plausibility. They note that it is the latter two criteria on which medically applied \xai generally fails and recommend that truthfulness is assessed using synthetic data, with known ground truth signals, where possible. This speaks to the additional rigors which any tool deployed in a critical setting, such as medical diagnostics, must meet.

\section{\specrex}

\begin{figure*}[t]
 \centering
  \includegraphics[width=\textwidth]{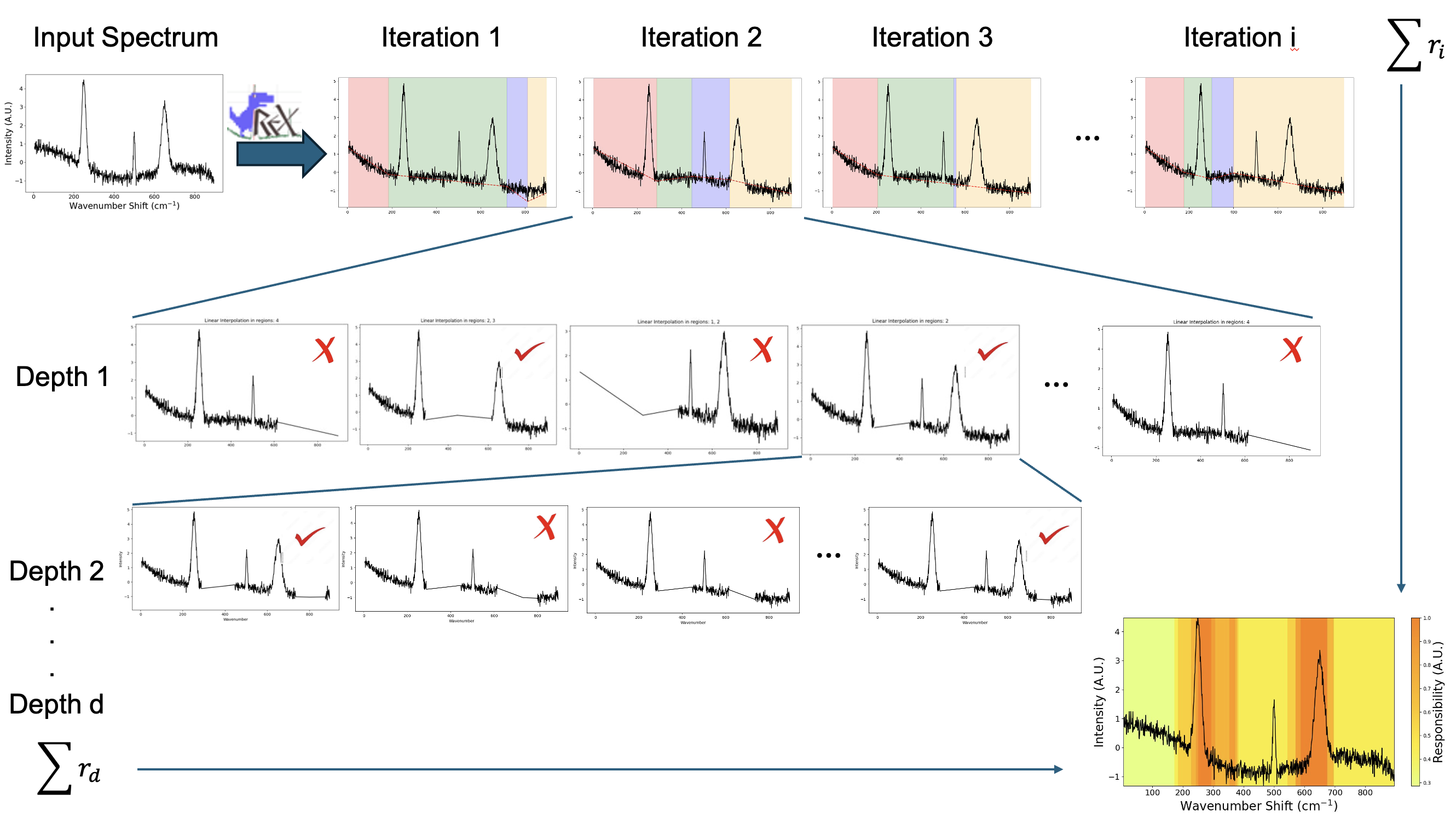}
  \caption{Schematic representation of \specrex. First, \specrex creates a set of mutants by generating random coordinates from the start to the end of the input wavenumber region, which are considered to be the initial splitting positions of the spectrum. The value from the previous split coordinate (start of the spectrum in the case of the first coordinate) is retained, while the rest of the spectrum is linearly interpolated. The model is then called for all the generated mutants. From the set of correctly classified mutants, the mutant with the smallest retained region is further explored, by recursively repeating the described procedure until the maximum search tree depth is reached, none of the mutants are classified correctly, or all generated mutants are too small. At each level of the search tree, to further isolate important regions, the start and end regions are set to the retained regions of the previous level, whilst the rest of the spectrum is inherited from the parent.}
  \label{fig:spectrum_generation}
\end{figure*}

\specrex is a bespoke \xai tool for RS (\Cref{fig:spectrum_generation}). We demonstrate proof-of-concept by creating a simulated dataset on which DL and \xai can be applied, allowing us to compare the explanations to the known ground truth. \specrex is an extension of the existing \xai tool for Causal \underline{Re}sponsibility-based E\underline{x}planations (\rex). First applied to explaining general image classification tasks \cite{chockler2024causal}, it has since been applied to medical images \cite{blake2023mrxai}. \rex uses causal responsibility \cite{chockler2004responsibility}, to provide a landscape from which an approximation of a minimal sufficient explanation is extracted. It works by iteratively refining occlusions over an image. Each time an occlusion is applied, the partially occluded image (referred to as a mutant) is used to query the ML model to see whether the occlusion changes the classification. If an occlusion does not change the classification, then the approximate degree of responsibility for the set of occlusions is distributed over the image pixels. Repeated applications of the algorithm allows a responsibility map to be built of the entire image, each pixel being ranked in terms of how responsible it is for the model's classification. From this map, \rex greedily extracts a minimal portion of the image which causes the model to classify the image to its class. 

\specrex extends this tool to be applicable to spectra. In images, occlusions are normally applied by zeroing out the pixels (or taking the mean or median value). However, in RS, this is indicative of a charge-coupled device saturated region. Instead, \specrex interpolates between the two endpoints of an occluded region, providing a plausible null region: that is, a region devoid of Raman features, but also devoid of induced features, such as saturation. \specrex by design seeks to make ``neutral'' occlusions to a spectrum. \specrex is model agnostic in the sense that it can be used on any ML model. It needs only to be able to run inference on a model and access the outputs. To make the occluded regions more natural, in the sense of bringing them closer to real-world RS data, 
a small amount of Gaussian noise is applied to the data interpolation.

\subsection{Raman Data Simulation and Model Development}

One of the problems with establishing \xai as a useful adjunct for biomedical DL applications is that its outputs (separate from the ML model's outputs) are hard to assess. This usually requires some form of ground truth, which is notoriously difficult to establish in biomedical studies. In order to demonstrate the potential usefulness of \xai in RS studies and to assess the fidelity of \specrex, we create a series of three increasingly complex simulated datasets.

\begin{figure}[t]
\centering
   \includegraphics[width=\columnwidth]{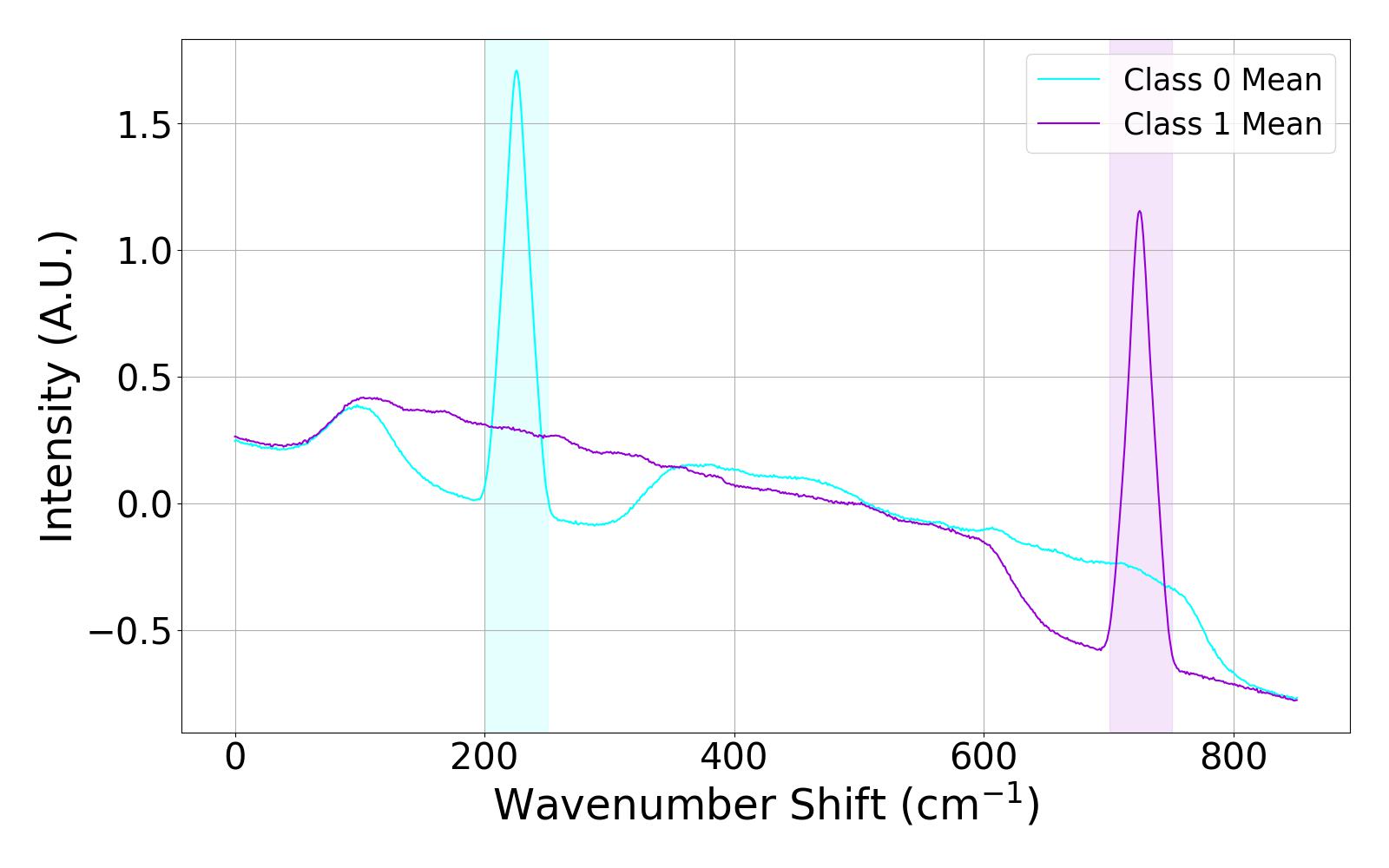}
   \caption{Mean spectra for the single peak dataset. The discriminating peaks for the focus classes are at 250 $cm^{-1}$ and 750 $cm^{-1}$, defining classes 0 and 1 respectively. The shaded regions highlight the discriminating features between classes.}
  \label{fig:SinglePeakMeans}
\end{figure}

\begin{figure}[t]
\centering
   \includegraphics[width=\columnwidth]{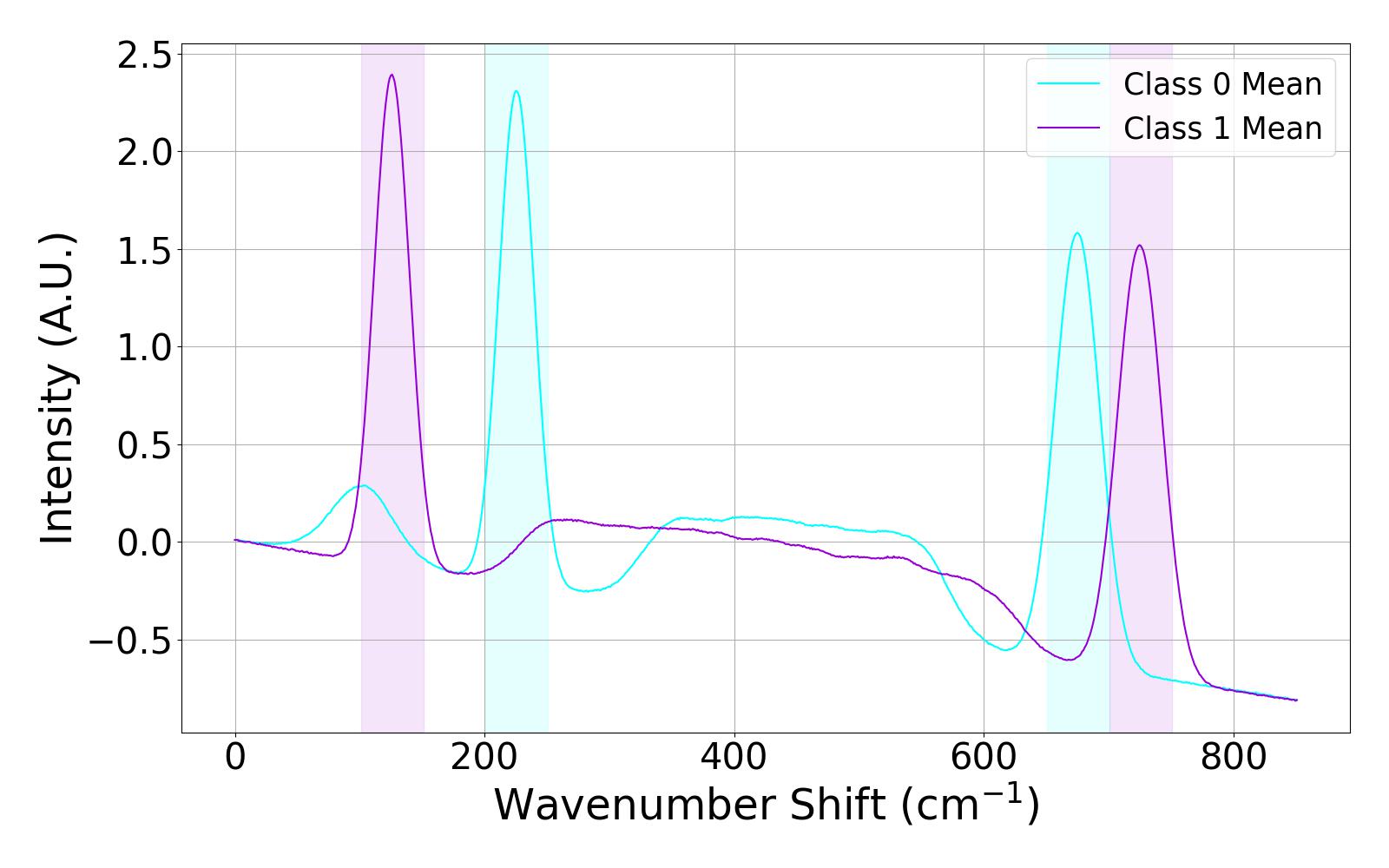}
   \caption{Mean spectra for the double peak dataset. The peaks at 250 $cm^{-1}$ and 750 $cm^{-1}$ are the discriminating peaks for class 0, while the peaks at 150 $cm^{-1}$ and 750 $cm^{-1}$ are the discriminating peaks for class 1. The shaded regions highlight the discriminating features between classes.}
  \label{fig:DoublePeakMeans}
\end{figure}

\begin{figure}[t]
\centering
   \includegraphics[width=\columnwidth]{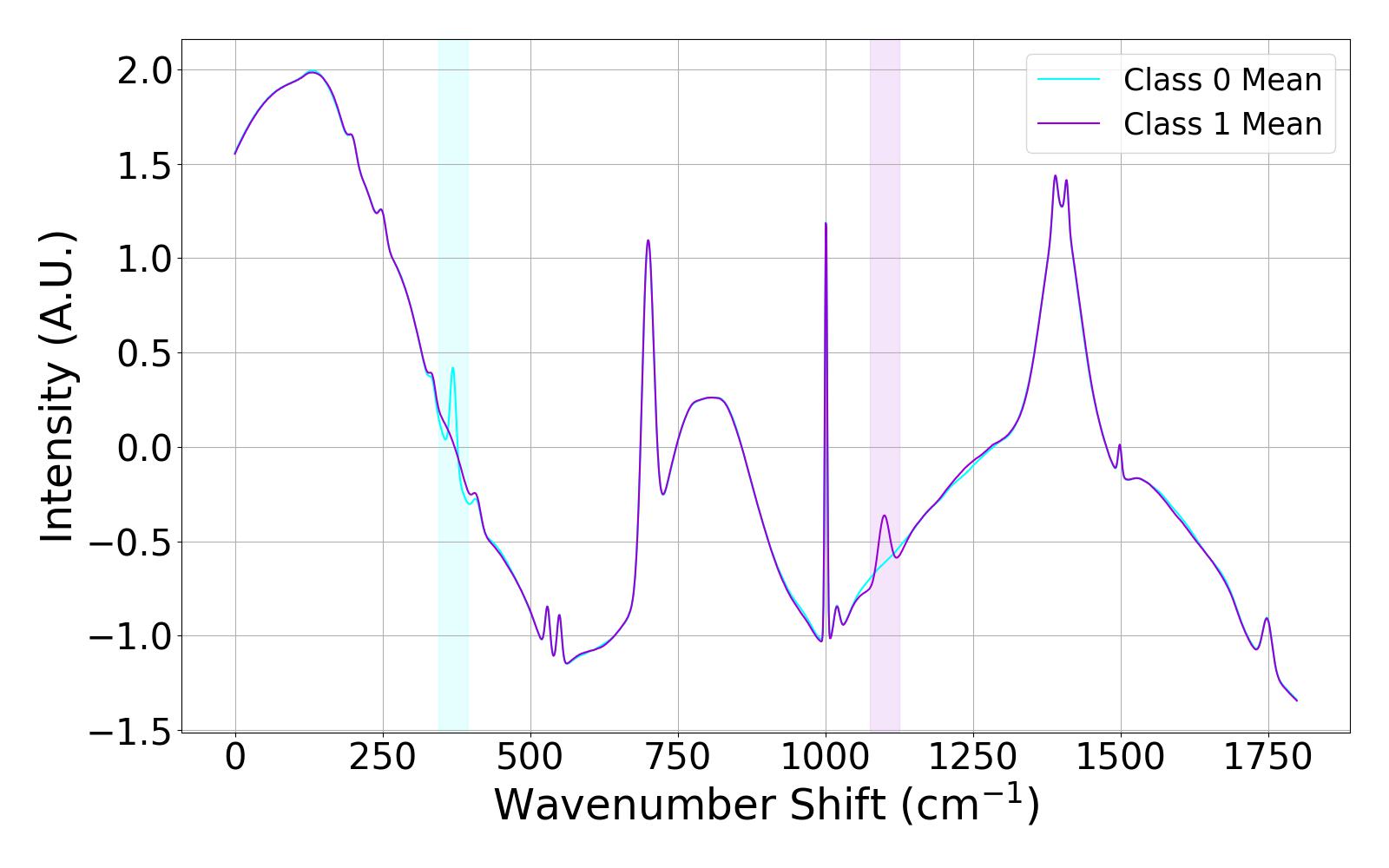}
   \caption{Mean spectra for the complex peak dataset. Each class has 18 Raman bands of varying breadth. The peaks at 370 $cm^{-1}$ and 1100 $cm^{-1}$ are the discriminating peaks for class 0 and class 1 respectively. The shaded regions highlight the discriminating features between classes.}
   \label{fig:ComplexMeanSpectra}
\end{figure}

To generate these simulated spectra, we randomly select positions for the required number of baseline points (we set it to 5 for all 3 datasets) and their corresponding mapping. These points act as anchors over which we fit a polynomial (cubic spline) in
order to get a spectral generation function. We then query this generation function over our desired wavenumber regions in order to get the skeleton of the simulated spectra. Peaks values are then calculated at the provided peak positions. To generate these, we apply 
a scaling factor (SF) based on the provided heights of the peaks using the formula:

\[
SF = \frac{H}{\sqrt{2 \pi W}}
\]

where H and W are the corresponding height and width of the peak. We then generate the peak by scaling the Gaussian Probability Density Function at the desired position by the scaling factor:

\[
\text{peak} = \frac{SF}{\sigma \sqrt{2 \pi}} \cdot e^{-\frac{1}{2} \left(\frac{x - \mu}{\sigma}\right)^2}
\]

Gaussian noise is introduced in order to simulate the noise present in real Raman spectra. 
This noise is then scaled with a noise scaling factor in order to control the amount of induced noise.
The final spectrum is the combination of the fitted polynomial, the peak positions, and the generated noise. For training, 10000 simulated spectra were created per class and for testing and \xai analysis, 500 per class.

In the simplest dataset, the first two classes consist of a single peak at 250 $cm^{-1}$ and 750 $cm^{-1}$, respectively, with a faux peak being placed at a random position in the spectra. The fixed peaks are the only discriminating features between the classes, while the faux peak is introduced in order to force the model to make a distinction between the important peaks and irrelevant features (\Cref{fig:SinglePeakMeans}). This dataset is used to test whether the \specrex explanation contains only the single defining peak in the simplest scenario. 

The second dataset is identical, except that each class has two distinguishing peaks. For class 0 these are at 250 $cm^{-1}$ and 750 $cm^{-1}$, and for class 1 at 150 $cm^{-1}$ and 650 $cm^{-1}$. The presence of both peaks is required for correct classification (\Cref{fig:DoublePeakMeans}). This dataset assesses the ability of \specrex to identify explanations of multiple spectral features that are relevant for classification, which may be distant on the wavenumber axis.

The third dataset is the most complex, with each class having $18$ spectral features, only one of which is discriminating: 370 $cm^{-1}$ for class 0 and 1100 $cm^{-1}$ for class 1 (\Cref{fig:ComplexMeanSpectra}). In addition, more noise was added. This represents a more complex dataset, closer to Raman spectra taken from biological sources compared to the previous simulated datasets.

Each dataset also has  a third ``everything-else'' class (not shown). This consists of examples where one or all of the distinguishing features from the first two classes are absent. This class is introduced to further induce the model to learn that the fixed peaks are the only distinguishing factors for classification and not the randomly generated peaks. This reduces the likelihood of the model learning spurious features in the simulated dataset, ensuring it learns the ground truth signal. Although these simulated datasets do not capture the full complexity of real Raman spectra, they are advantageous for a proof-of-concept study, as the ground truth for them is known and hence
can be used to assess \specrex.

To classify these datasets a series of CNN + LSTM (Long Short-Term Memory) models, also known as Long-term Recurrent Convolutional Networks (LRCNs), were developed, each with 4 convolutional layers, 3-layer LSTMs and two fully connected layers. This model configuration was chosen in order to capture spatial dependencies of real, and of our simulated, Raman spectra. It is important to note that the DL model itself is not the subject of scrutiny of this paper. We chose a LRCN model as it is representative of DL models, which are notoriously the most opaque of ML models. No attempt was made to optimize these models. The models were able to achieve 100\% accuracy on the double peak dataset, and approximately 83\% accuracy on the single and complex peak datasets. Although this is an unrealistically simplified model development pipeline, it affords control of extraneous variables, allowing us to assess \specrex in isolation.

\section{Results}

There are no generally agreed upon metrics to quantitatively measure the quality of explanations. In the biomedical context, an explanation is as good as a clinician feels it facilitates decision making. In lieu of such data (via a user study) a number of quantitative measures do exist, such as Hausdorff distances or Dice Coefficients. However, in biomedical RS datasets, it is not clear which, if any, method is most suitable. We employ a very simple technique, whereby we count the number of peaks an explanation tool returns. This method captures the intuition that a clinical spectroscopist would need to disentangle which 'explanations' are true and which, if any, are useful.

Using this method, we compare \specrex with several variants of SHAP (Kernel, DeepLIFT and Gradient). We chose SHAP as a baseline comparison due to its enduring popularity, and its use in at least two biomedical RS studies \cite{bellantuono2023explainable,ibtehaz2023ramannet}. This is accompanied by qualitative comparisons of the \xai techniques from selected spectra. To facilitate this comparison the \specrex and SHAP outputs have been modified to a common format.

\begin{table}[t]
\caption{Number of Peaks for Class 0}
\centering
\begin{tabular}{lcccc}
\toprule
Dataset       & SpecReX & KernelSHAP & GradientSHAP & DeepLIFTSHAP \\
\midrule
Single Peak   & 2.2     & 284.84     & 281.92       & 276.46       \\
Double Peak   & 22.62   & 281.52     & 281.38       & 279.38       \\
Complex Peak  & 1.06    & 600.56     & 582.14       & 582.82       \\
\bottomrule
\end{tabular}
\label{tab:class_0_peakcount}
\end{table}

\begin{table}[t]
\caption{Number of Peaks for Class 1}
\centering
\begin{tabular}{lcccc}
\toprule
Dataset       & SpecReX & KernelSHAP & GradientSHAP & DeepLIFTSHAP \\
\midrule
Single Peak   & 1.78    & 282.6      & 280.62       & 277.38       \\
Double Peak   & 5.46    & 282.62     & 281.14       & 279.04       \\
Complex Peak  & 10.66   & 599.34     & 581.66       & 583.66       \\
\bottomrule
\end{tabular}
\label{tab:class_1_peakcount}
\end{table}

\begin{figure*}[htbp]
    \centering
    \begin{subfigure}{\textwidth}
        \centering
        \includegraphics[width=\textwidth]{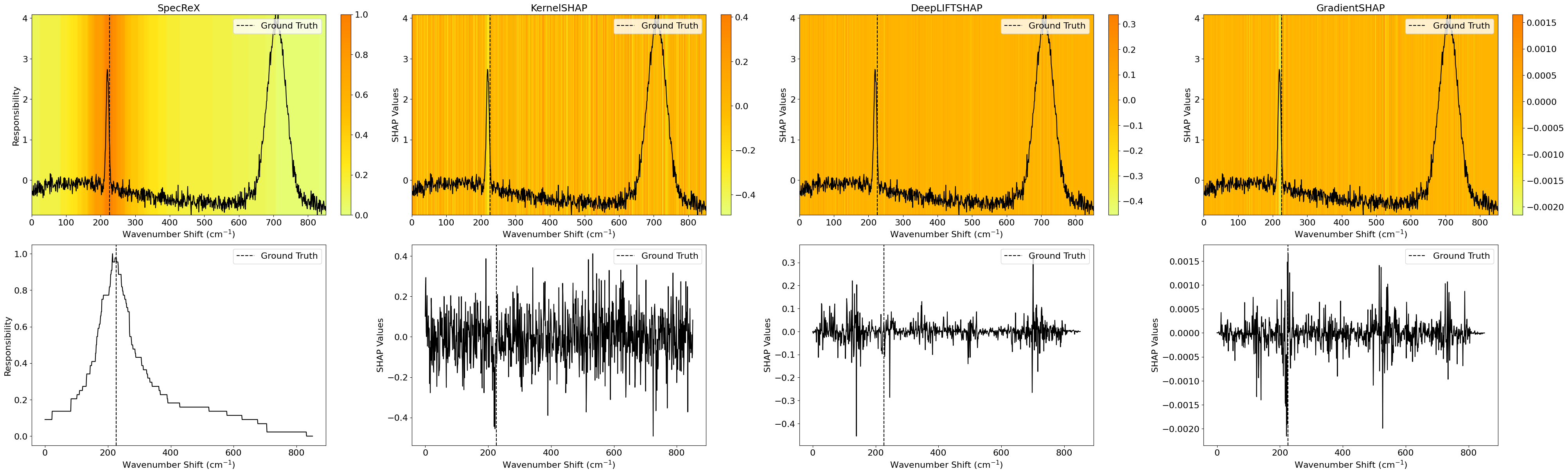}
        \caption{Single Peak Dataset}
        \label{fig:class_0_plot_single}
    \end{subfigure}
    
    \vspace{0.5cm} 
    
    \begin{subfigure}{\textwidth}
        \centering
        \includegraphics[width=\textwidth]{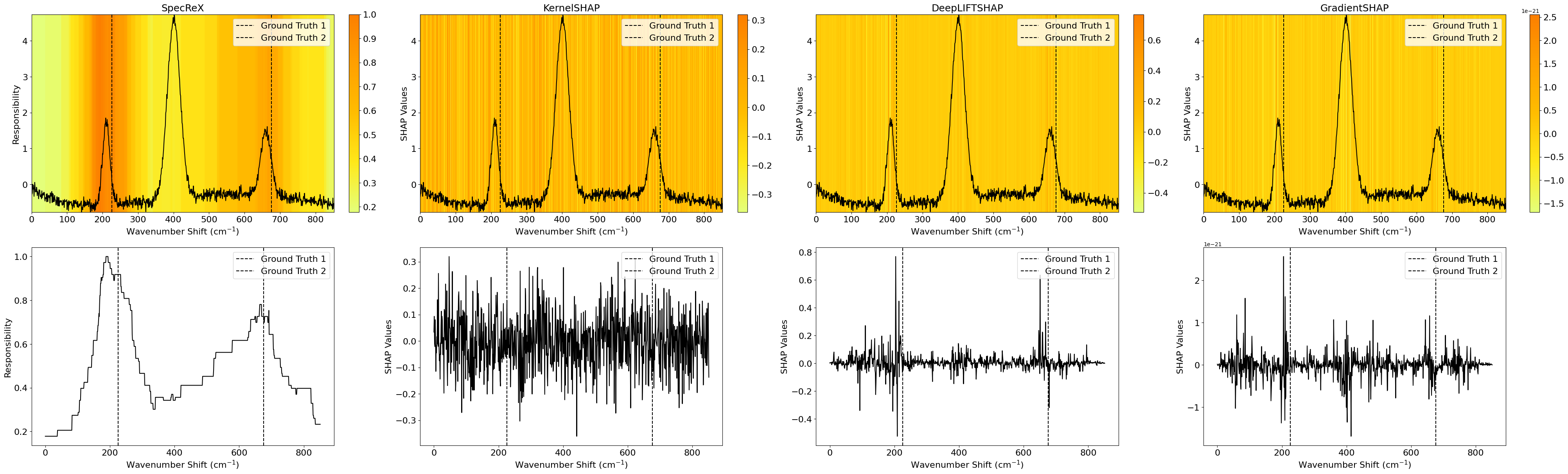}
        \caption{Double Peak Dataset}
        \label{fig:class_0_plot_double}
    \end{subfigure}
    
    \vspace{0.5cm} 
    
    \begin{subfigure}{\textwidth}
        \centering
        \includegraphics[width=\textwidth]{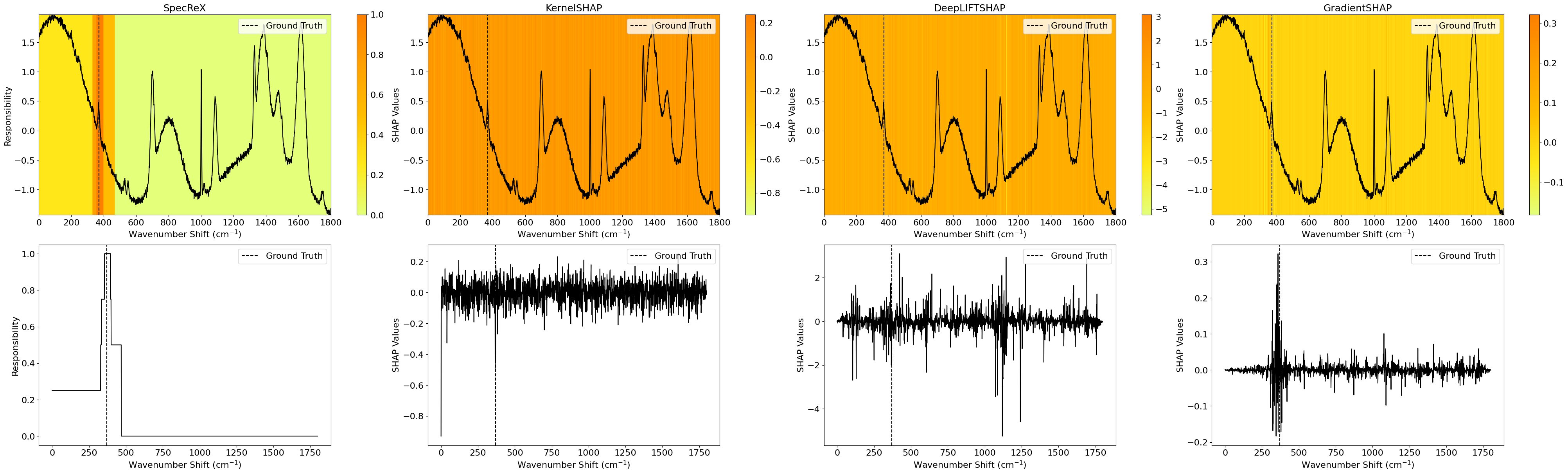}
        \caption{Complex Peak Dataset}
        \label{fig:class_0_plot_complex}
    \end{subfigure}
    
    \caption{Class 0 results. Responsibility/saliency maps of selected spectra with plots of map values below. Leftmost is \specrex, followed by KernelSHAP, DeepLIFTSHAP and with GradientSHAP rightmost. Top row, single peak; middle row, double peak and bottom row, complex peak. Black dashed vertical line is the ground truth peak location. Of note is relative homogeneity of SHAP explanations compared to \specrex, which draws to the eye to relevant features with far less noise.}
    \label{fig:class_0_results}
\end{figure*}

\begin{figure*}[htbp]
    \centering
    \begin{subfigure}{\textwidth}
        \centering
        \includegraphics[width=\textwidth]{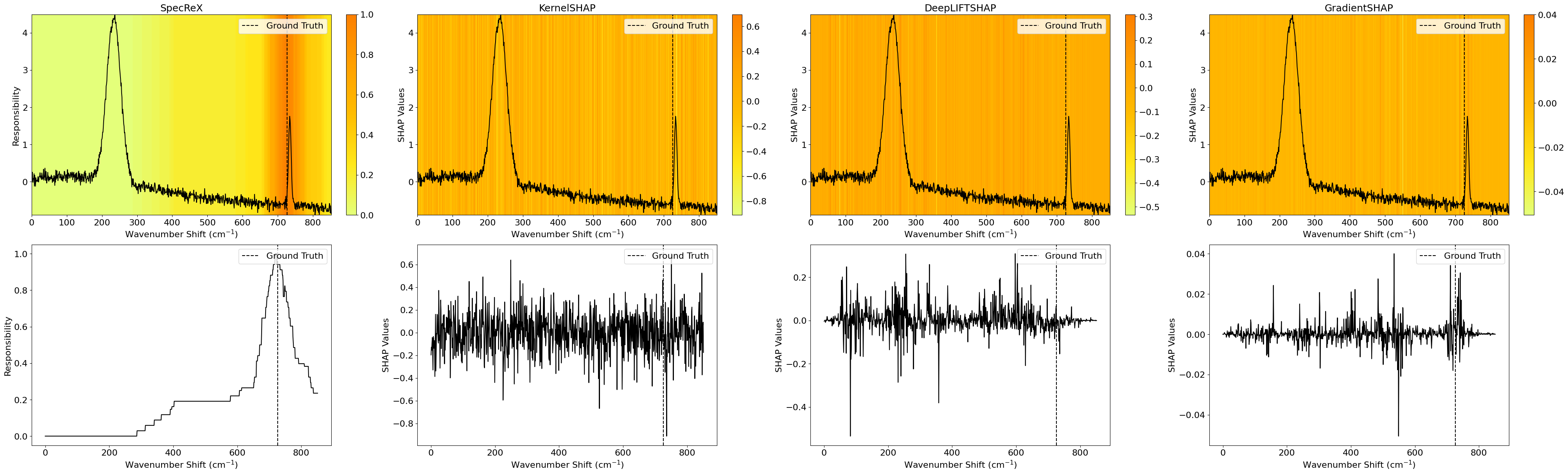}
        \caption{Single Peak Dataset}
        \label{fig:class_1_plot_single}
    \end{subfigure}
    
    \vspace{0.5cm} 
    
    \begin{subfigure}{\textwidth}
        \centering
        \includegraphics[width=\textwidth]{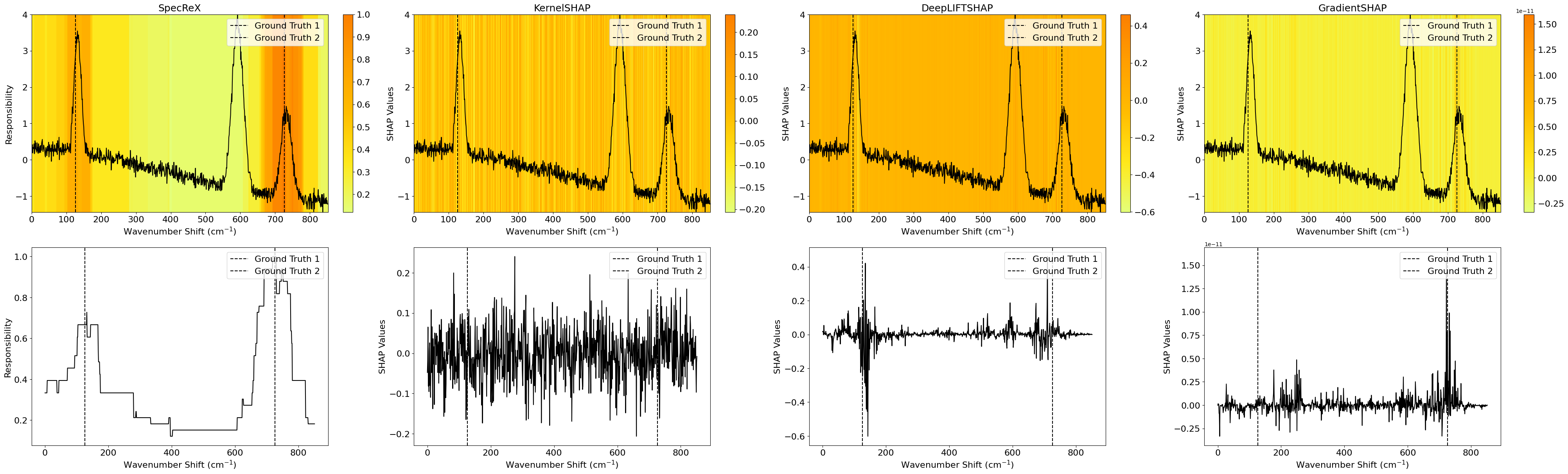}
        \caption{Double Peak Dataset}
        \label{fig:class_1_plot_double}
    \end{subfigure}
    
    \vspace{0.5cm} 
    
    \begin{subfigure}{\textwidth}
        \centering
        \includegraphics[width=\textwidth]{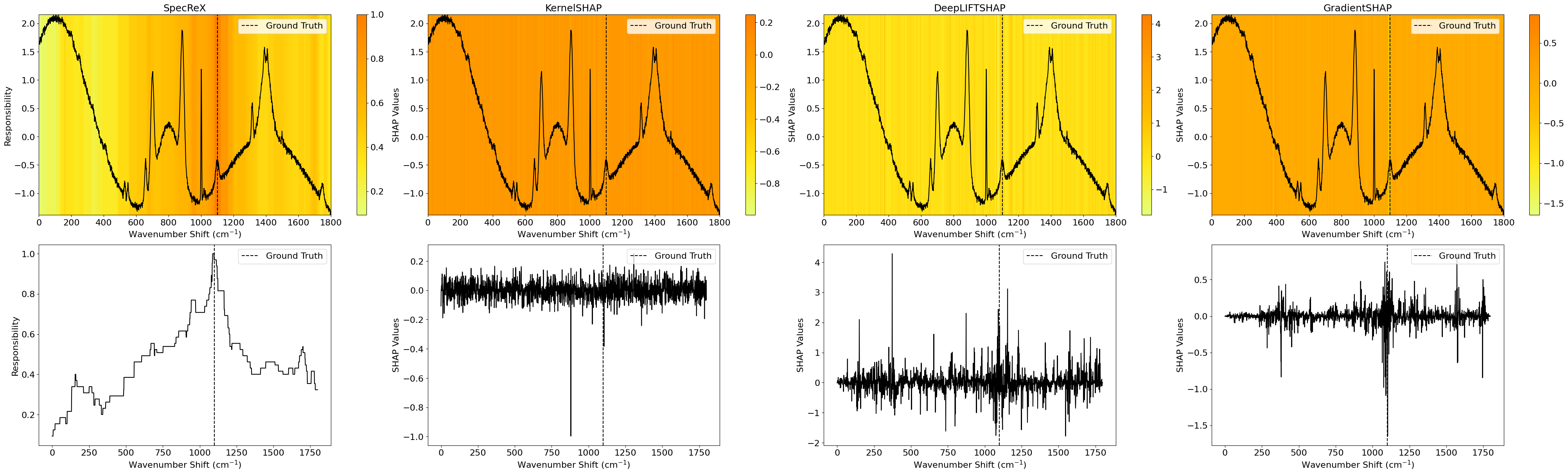}
        \caption{Complex Peak Dataset}
        \label{fig:class_1_plot_complex}
    \end{subfigure}
    
    \caption{Class 1 results. Responsibility/saliency maps of selected spectra with plots of map values below. Leftmost is \specrex, followed by KernelSHAP, DeepLIFTSHAP and with GradientSHAP rightmost. Top row, single peak; middle row, double peak and bottom row, complex peak. Black dashed vertical line is the ground truth peak location.}
    \label{fig:class_1_results}
\end{figure*}

\subsection{Single Peak}

This dataset has only a single distinguishing peak for the two relevant classes, the other peak being spurious: the model should ignore it. \Cref{fig:class_0_plot_single} and \Cref{fig:class_1_plot_single} demonstrate that \specrex is able to show that the model is using the ground truth peak to classify the spectrum correctly while ignoring the spurious peak. It also shows that while the SHAP variants can also localize to the ground truth, this is accompanied by a high degree of noise sufficient to obscure interpretation. This manifests in table \Cref{tab:class_0_peakcount} which shows that the SHAP variants return significantly more peaks than \specrex. This is also true of the class 1 dataset, seen in table \Cref{tab:class_1_peakcount}.

\subsection{Double Peak}

Both of the classes in this dataset contains two distinguishing peaks, again with an uninformative third peak added at a random position. The occurrence of both of the distinguishing peaks are necessary to correctly classify the spectra into one of the two classes. The purpose of this simulated dataset is to show what happens when multiple spatially distant but necessary peaks are present, as is common in spectra from biological sources. \Cref{fig:class_0_plot_double} and \Cref{fig:class_1_plot_double} show that indeed, both diagnostic peaks have been highlighted by \specrex, although one more so than the other. Similar to the single peak dataset, even those SHAP variants which do localize to the ground truth peaks are accompanied by significant noise.

\subsection{Complex Peak}

This dataset was constructed to be at least somewhat representative of the complex nature of Raman spectra taken from biological sources. Both classes are distinguished by only a single peak. As shown in \Cref{fig:class_0_plot_complex} and \Cref{fig:class_1_plot_complex} \specrex localizes to the sole discriminating spectral region ignoring the spurious peaks. Again, the interpretation of the SHAP variants is a little more fraught.

\section{Discussion}

 We see in the explanations generated by \specrex and the SHAP variants, that \specrex provides a more targeted explanation. \specrex has a set scale (responsibility values are always bounded between 0 and 1), which is easily interpretable and can be compared in a direct manner across explanations from multiple examples. Conversely, SHAP values are unbounded and can result in values that are drastically different (and can also be negative even within the ground truth), making it difficult to intuitively interpret. Furthermore, the unbounded nature of the SHAP values makes it difficult to compare explanations across multiple examples.

This demonstrates that \specrex is able to provide explanations which locate to the discriminating features of increasingly complicated simulated spectra with known ground truths. Most importantly, \specrex does not provide ``noisy'' explanations, even on those models with relatively low accuracy (83\%). This serves as a sanity check, ensuring that the \xai tool is able to work as claimed.

This helps address two of the criteria Jin \textit{et al.} cite as important for assessing medical \xai for \cite{jin2023guidelines}: ``truthfulness'' and, partially, ``informative plausibility''. The former has been explicitly addressed based on their recommendation of assessing against a ``synthetic dataset with known discriminative features''. This criterion is also a prerequisite for ``informative plausibility'', in which the plausibility of the explanation is judged by domain experts. This cannot be fully satisfied with a simulated dataset, but it does demonstrate how this could be explored. By identifying those features which the model uses to discriminate between classes, it facilitates a guided exploration of the biochemical antecedents of the identified spectral features. This can achieved by correlating the identified spectral features with those associated with a specific disease process. Once trust in both the model and the \xai tool has been vindicated, attention may shift to hypothesis generation: allowing domain experts to probe hitherto unknown spectral features associated with a disease process.

This is predicated upon the model (and training process) being ``good''. The medical literature is replete with examples of models with excellent performance, as measured with traditional metrics such as F1 scores and receiver operating curves, but failing to generalize to the clinical setting \cite{hutson2018artificial,mcdermott2021reproducibility,volovici2022steps}. This is sometimes referred to as the generalization gap. This can manifest with \xai as unexpected and spurious features being highlighted (as was the case with the conflation of eyeballs and tumors in models of brain MRI data). Thus, \xai facilitates domain experts to carefully examine the model outputs to determine whether it is identifying plausible features or has overfit to spurious features.

\subsection{Future Development and Validation}

\specrex exists as a proof-of-concept and remains under active development. Future work will involve two primary directions: further tool development to increase its fidelity and more rigorous testing to ensure its validity to clinical problems. A key aspect of \specrex is that it is guided by domain specific knowledge to generate its occlusions, thus incorporating inductive biases. This differs from other \xai tools, which use generic occlusion methods. In future work, we will further exploit this ability by adding non-linear interpolations for occluded regions and adding Poisson noise (as RS is ultimately beholden to shot noise) to these regions. This will make for more realistic, smoothly varying low information regions. We can then assess whether these more realistic occlusions, for instance Poisson rather than Gaussian noise, lead to more accurate explanations.

We will also focus on more extended validations of \specrex.  We will more thoroughly assess the validity of \specrex by creating an \textit{in vitro} ground truth using biomolecules combined under laboratory conditions and then subject to RS. A difficulty in assessing the validity of an \xai tool is in establishing whether the DL model has learned spurious features or if the \xai tool is erroneously identifying regions. The \textit{in vitro} dataset will allow us to explore how a model trained upon real data interacts with \specrex and assess to what degree both the model and \specrex are able to locate the ``ground truth'' signal. However, this is still an unrealistically controlled situation compared to clinical applications. The following step will be to apply \specrex to \textit{ex vivo} RS data taken from clinically motivated oncology studies. This has the disadvantage of having an ambiguous ground truth. In lieu of a hard ground truth, we will discuss with leading histopathologists as to the plausibility of the explanations in the context of the known pathogenesis, and any putative biochemical antecedents, of pertinent disease classes.

Our discussions with clinicians will also extend to theoretical considerations which may impact the ultimate usefulness of an explanation. There are several over-lapping and competing definitions of an explanation in the \xai literature. In this paper we have used a definition based on the degree of responsibility of each input feature to the classification. However, the philosophical and mathematical underpinnings of actual causality, upon which \rex is based, has a more formal definition of an explanation as a minimal set of features required to retain a model's classification. In future work we will explore whether this definition allows for more useful explanations from the perspective of clinical decision making, which ultimately must be the barometer of success for any clinically deployed \xai tool.

\section{Conclusion}

We have explained the need for \xai applied to biomedical RS,  outlined the unique challenges faced by such tools when deployed in critical settings such as medical diagnostics and introduced and compared the validity of one such tool, \specrex, to some SHAP variants upon simulated Raman data. Under idealized conditions we have established that \specrex is able to localize the spectral features which cause a DL model to correctly classify spectra. As RS becomes further embedded in the AI ecosystem it will increasingly be beholden to regulatory, clinical and scientific requirements, some of which can be satisfied with \xai.

By providing an explanation of what spectral features a model is using to classify between disease states, \specrex could help satisfy the coming regulatory requirements that any AI  (especially those with diagnostic capabilities) is transparent enough so that its outputs can be explained. Although this paper has been motivated in terms of biomedical applications, the approach is general and could be extended to any domain of vibrational spectroscopy in which \xai would be helpful.




\section*{Funding} Blake, Kelly, Chockler and Chanchal were supported in part by the UKRI Trustworthy Autonomous Systems Hub (EP/V00784X/1), the UKRI Strategic Priorities Fund to the UKRI Research Node on Trustworthy Autonomous Systems Governance and Regulation (EP/V026607/1), and CHAI - EPSRC AI Hub for Causality in Healthcare AI with Real Data (EP/Y028856/1).

\bibliographystyle{splncs04}
\bibliography{rsc}
\end{document}